\documentclass[10pt,twocolumn,letterpaper]{article}

\usepackage{iccv}
\usepackage{times}
\usepackage{epsfig}
\usepackage{graphicx}
\usepackage{amsmath}
\usepackage{amssymb}
\usepackage{multirow}
\usepackage{bbding}
\usepackage{pifont}
\usepackage{fontawesome}
\usepackage{bm}
\usepackage{colortbl}
\usepackage[accsupp]{axessibility} 
\usepackage[labelformat=simple]{subcaption}
\definecolor{graycolor}{rgb}{0.95,0.95,0.95}

\usepackage[pagebackref=true,breaklinks=true,letterpaper=true,colorlinks,bookmarks=false]{hyperref}
\iccvfinalcopy 

\ificcvfinal\fi

\begin{document}

\title{MPI-Flow: Learning Realistic Optical Flow with Multiplane Images}

\author{Yingping Liang\textsuperscript{$1$} \qquad \qquad Jiaming Liu\textsuperscript{$2$} \qquad \qquad Debing Zhang\textsuperscript{$2$} \qquad \qquad  Ying Fu\textsuperscript{$1$}\thanks{Corresponding Author: fuying@bit.edu.cn} \\
	\textsuperscript{$1$}Beijing Institute of Technology
 \quad \textsuperscript{$2$}Xiaohongshu Inc. \\ 
}

\maketitle
\ificcvfinal\thispagestyle{empty}\fi
\begin{abstract}
The accuracy of learning-based optical flow estimation models heavily relies on the realism of the training datasets. Current approaches for generating such datasets either employ synthetic data or generate images with limited realism. However, the domain gap of these data with real-world scenes constrains the generalization of the trained model to real-world applications. To address this issue, we investigate generating realistic optical flow datasets from real-world images. Firstly, to generate highly realistic new images, we construct a layered depth representation, known as multiplane images (MPI), from single-view images. This allows us to generate novel view images that are highly realistic. To generate optical flow maps that correspond accurately to the new image, we calculate the optical flows of each plane using the camera matrix and plane depths. We then project these layered optical flows into the output optical flow map with volume rendering. Secondly, to ensure the realism of motion, we present an independent object motion module that can separate the camera and dynamic object motion in MPI. This module addresses the deficiency in MPI-based single-view methods, where optical flow is generated only by camera motion and does not account for any object movement. We additionally devise a depth-aware inpainting module to merge new images with dynamic objects and address unnatural motion occlusions. We show the superior performance of our method through extensive experiments on real-world datasets. Moreover, our approach achieves state-of-the-art performance in both unsupervised and supervised training of learning-based models. The code will be made publicly available at: \url{https://github.com/Sharpiless/MPI-Flow}.
\end{abstract}

\begin{figure}[t]
   \centering
    \subcaptionbox{\label{FG1-dDAVIS} Depthstill. \cite{aleotti2021learning}}{\includegraphics[width = .32\linewidth]{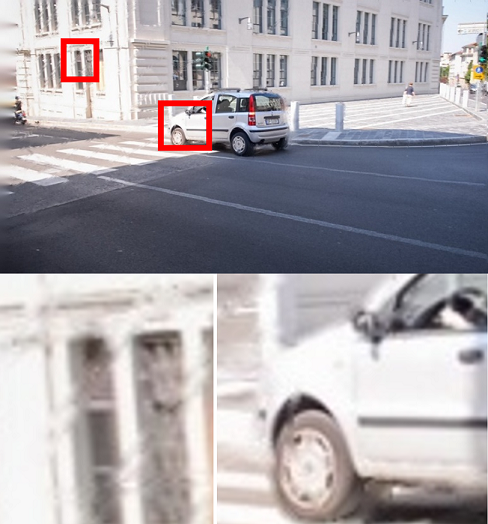}}\hfill
    \subcaptionbox{\label{FG1-RF-DAVIS} RealFlow \cite{han2022realflow}}{\includegraphics[width = .32\linewidth]{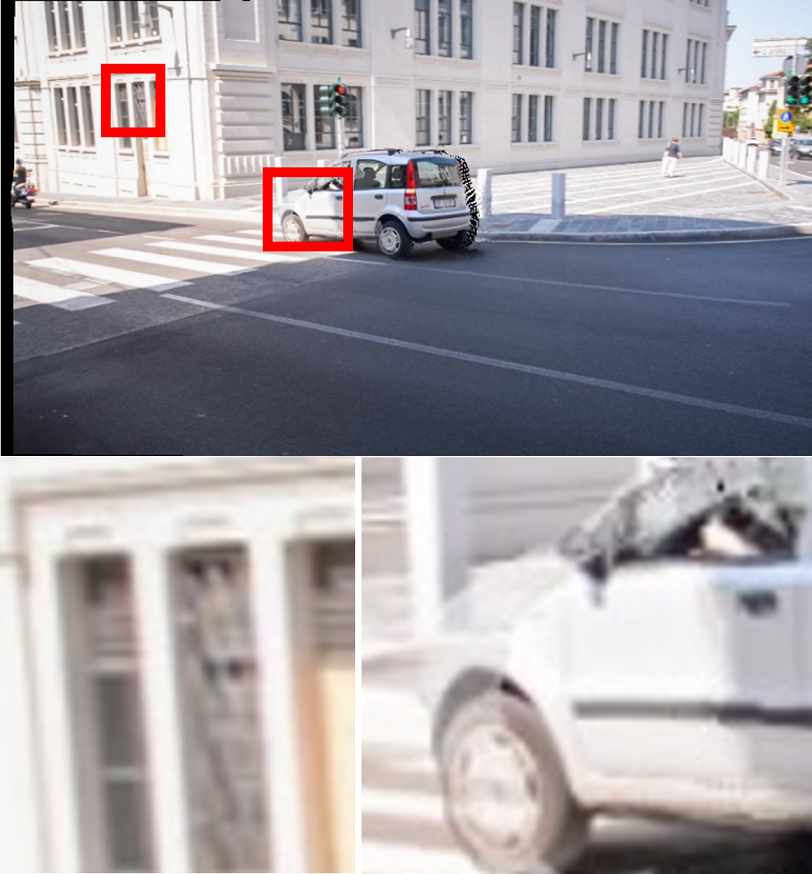}}\hfill
    \subcaptionbox{\label{FG1-MF-DAVIS} Ours}{\includegraphics[width = .32\linewidth]{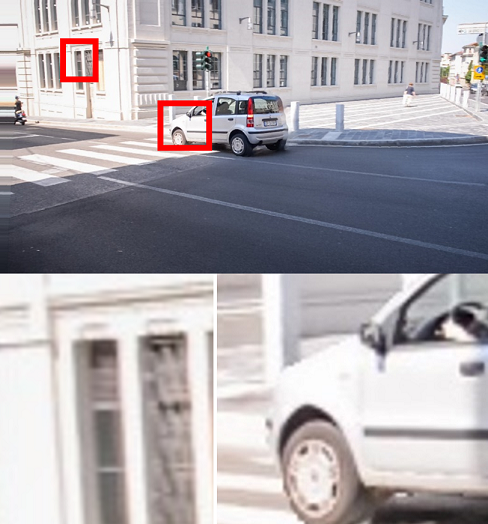}}\hfill
    \subcaptionbox{\label{FG1-flow} Source image and details of generated flows from our method}{\includegraphics[width = .99\linewidth]{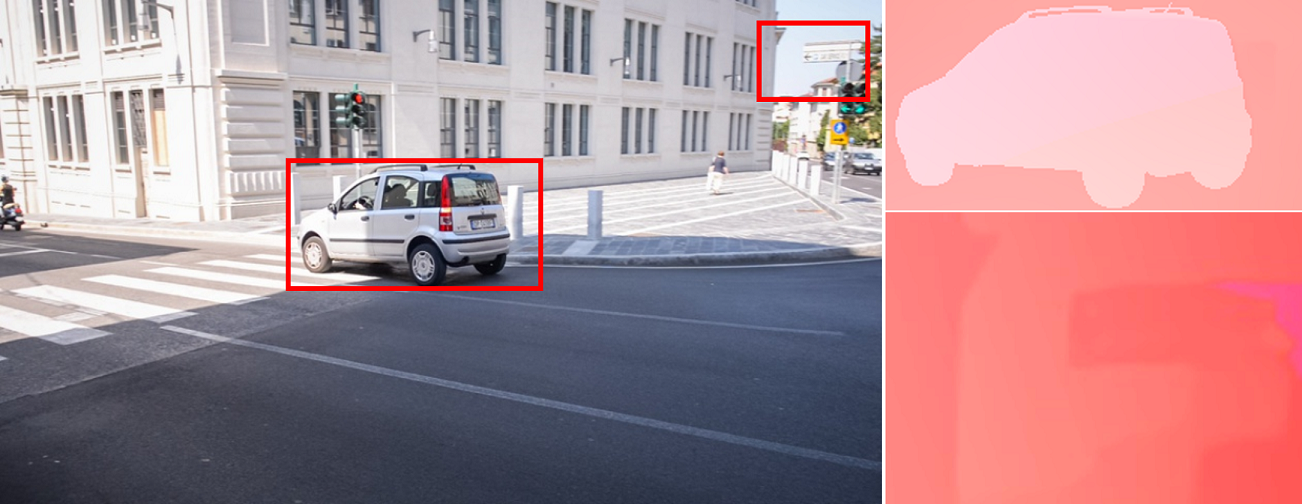}}\hfill
    \caption{Visual comparison of images reveals that our method, when compared with those based on real-world images \cite{aleotti2021learning, han2022realflow}, demonstrates superior image realism. Our method also achieves fine optical flows by employing volume rendering in MPI and separate motions of the camera and object with the independent object motion module.}
    \label{fig:framework-v3}
\end{figure}

\section{Introduction}
Optical flow refers to the precise calculation of per-pixel motion between consecutive video frames. Its applications span a wide range of fields, including object tracking \cite{8735957, zhang2020multiple}, robot navigation \cite{karoly2019optical, sanket2021prgflow}, three-dimensional (3D) reconstruction \cite{kokkinos2021learning, hanari2022image}, and visual simultaneous localization and mapping (SLAM) \cite{zhang2020flowfusion, cheng2019improving, liu2021rdmo}. In recent years, with the rapid development of neural networks, learning-based methods \cite{sun2018pwc, teed2020raft} have demonstrated significant advances compared to traditional model-based algorithms \cite{brox2009large, weinzaepfel2013deepflow, zach2007duality}. Conventional practices primarily rely on synthetic data, as demonstrated by \cite{dosovitskiy2015flownet, ilg2017flownet, Butler:ECCV:2012}. Synthetic data contains exact optical flow labels and animated images. However, the domain gap between synthetic and real data hinders its further improvements in real-world applications. 

Recent studies have aimed to extract optical flow from real-world data by employing hand-made special hardware \cite{baker2011database, geiger2012we, menze2015object}. However, the rigidly controlled and inefficient collection procedure limits their applicability. To address this issue, Depthstillation \cite{aleotti2021learning}, and RealFlow \cite{han2022realflow} have been proposed, which project each pixel in the real-world image onto the novel view frame with the help of random motions of virtual cameras or estimated flows. Nonetheless, both methods are limited by the lack of image realism, leading to issues such as collisions, holes, and artifacts, as illustrated in Figure \ref{fig:framework-v3}. These limitations constrain the real-world performance of learning-based optical flow models \cite{sun2018pwc, teed2020raft}.

To achieve higher image realism, we turn our attention to the use of single-view Multiplane Images (MPI) \cite{zhou2018stereo, single_view_mpi, zhou2021cross, Carvalho2021LearningMI, han2022single}. This line of work demonstrates remarkable single-view image rendering capabilities and effectively reduces collisions, holes, and artifacts commonly found in previous methods \cite{aleotti2021learning, han2022realflow}. These advancements contribute to higher image realism, prompting a natural question: Can high-realistic MPI methods be adapted to generate high-quality optical flow datasets for training purposes?

To this end, we propose \textit{MPI-Flow}, aiming to generate realistic optical flow datasets from real-world images. Specifically, we first review the image synthesis pipeline of MPI and devise an optical flow generation pipeline along with image synthesis. In this step, we build an MPI by warping single-view image features onto each layered plane with the predicted color and density. The color and density then be mapped into a realistic new image via volume rendering. With the layered planes, we extract optical flows with virtual camera motions from the rendered image and the real image. Second, as the MPI can only be applied in static scenes, which yield limited motion realism, we propose an independent object motion module and a depth-aware inpainting module to tackle this issue. The independent object motion module decouples dynamic objects from static scenes and applies different virtual camera matrices to calculate the motion of both dynamic and static parts. The depth-aware inpainting module is introduced to remove the object occlusion in the synthesized new image.

With MPI-Flow, a large number of single-view images can be used to generate large-scale training datasets with realistic images and motions. This enables learning-based optical flow models better generalization to a wide range of real-world scenes. Extensive experiments on real datasets demonstrate the effectiveness of our approach. In summary, our main contributions are as follows:

\begin{itemize}
\item We are the first to present a novel MPI-based optical flow dataset generation framework, namely MPI-Flow, which can significantly improve the realism of the generated images and motion.
\item We present a novel independent object motion module for modeling dynamic objects in MPI, which can model realistic optical flow from camera motion and object motion simultaneously.
\item We design a depth-aware inpainting module for realistic image inpainting, which can remove unnatural motion occlusions in generated images.
\end{itemize}

\begin{figure*}[t]
\begin{center}
    \includegraphics[width=0.95\linewidth]{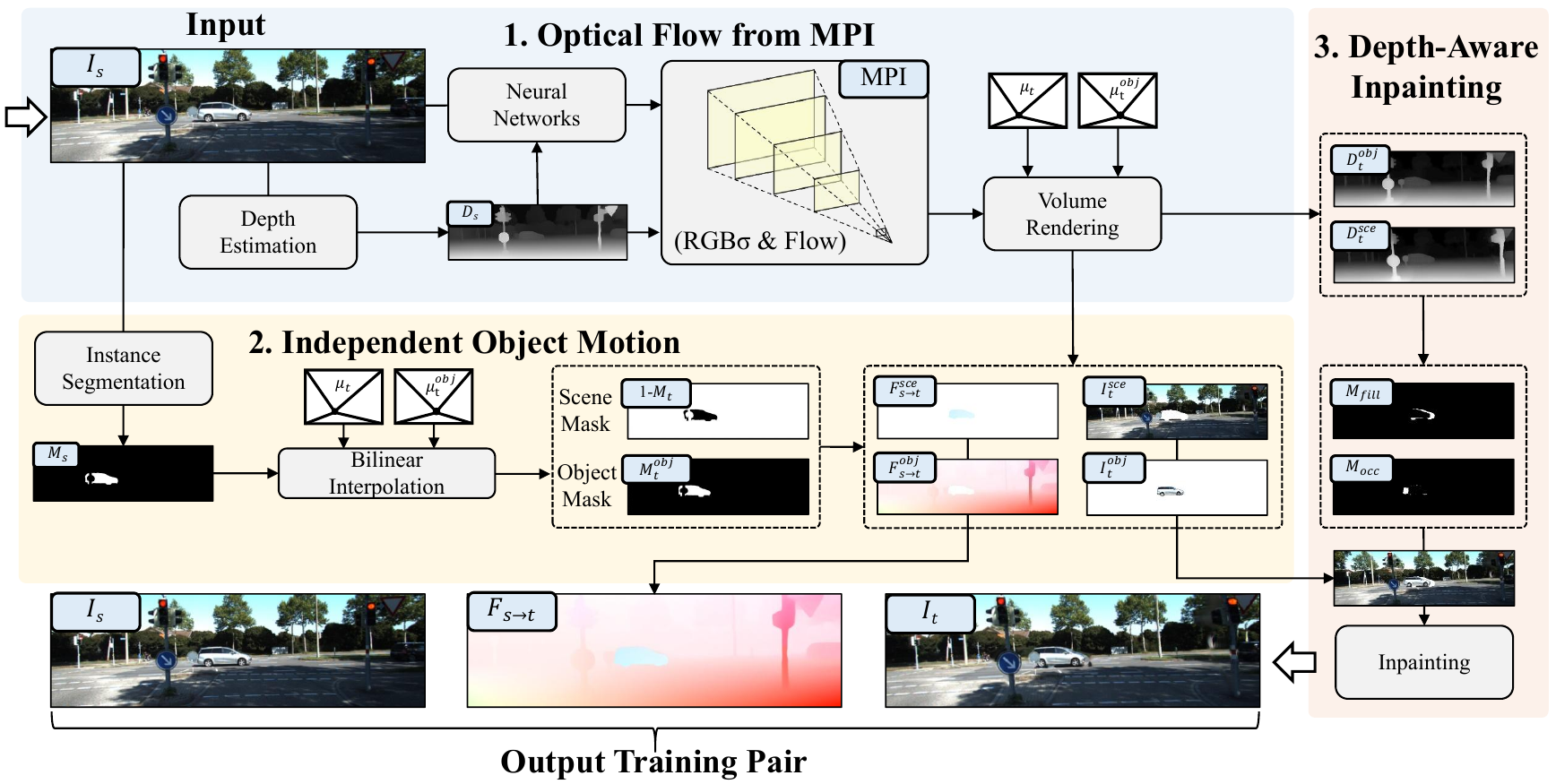}
    \end{center}
    \caption{Illustration of our proposed MPI-Flow framework with single-view image $\textbf{I}_{s}$ as input. We estimate depths to construct MPI where RGB and density of each plane are predicted by neural networks and the flow of each plane is calculated with camera matrixes. Both the novel views and flow maps are rendered by volume rendering and separated by the independent object motion module with novel view object masks. The new image is merged by depth-aware inpainting.}
    \label{fig:framework-MPI-Flow}
\end{figure*}

\section{Related Work}

In this section, we review the most relevant studies on optical flow networks, optical flow dataset generation, and novel view image synthesis methods.

\noindent \textbf{Supervised Optical Flow Network.} Early methods train deep neural networks to match patches across images \cite{weinzaepfel2013deepflow}. FlowNet \cite{dosovitskiy2015flownet} first trains convolutional neural networks on the synthetic datasets with optical flow. Moreover, the follow-up methods \cite{hui2018liteflownet, hur2019iterative, ilg2017flownet, luo2022learning, luo2022learning2} with advanced modules and network architectures make a significant improvement in supervised optical flow learning, with RAFT \cite{teed2020raft} representing state-of-the-art. However, generalization remains a cause for concern due to the domain gap between synthetic datasets and real-world applications. To address this problem, our work focuses on generating realistic optical flow datasets from real-world images.

\noindent \textbf{Dataset Generation for Optical Flow.} The use of fluorescent texture to record motions in real-world scenes is first described in \cite{baker2011database} to obtain flow maps. KITTI \cite{geiger2012we, menze2015object} provides sophisticated training data through complex lidar and camera setups. However, the aforementioned real-world datasets have limited quantities and constrained scenes, making it difficult for models trained using deep supervised learning to generalize to more expansive scenes. Synthesized training pairs, such as those in Flyingchairs \cite{dosovitskiy2015flownet} and Flyingthings \cite{ilg2017flownet}, have shown promise for supervised learning. However, moving animated image patches cannot accurately match real-world scenes, leading to domain gaps. AutoFlow \cite{sun2021autoflow} introduces a learning-based approach for generating training data by hyper-parameters searching. However, AutoFlow relies on optical flow labels for domain adaptation, which is not practical in most scenarios where ground truth labels are unavailable.

Two recent works have proposed methods for generating training datasets based on real-world images or videos. The first, called Depthstillation \cite{aleotti2021learning}, synthesizes paired images by estimating depth and optical flows from a single still image. Optical flows are calculated based on the virtual camera pose and depth. The second method, called RealFlow \cite{han2022realflow}, synthesizes intermediate frames between two frames using estimated optical flows with RAFT. However, both methods use naive image synthesis techniques that fail to meet the demand for realism criteria due to hole-filling and artifacts. In contrast, our method improves on this approach by using a well-designed and modified multiplane image (MPI) technique to obtain realistic images.

\noindent \textbf{Novel View Synthesis.} View synthesis methods aim to generate new images from arbitrary viewpoints by utilizing a given scene. Several classical approaches \cite{tulsiani2018layer, lombardi2019neural, zhou2018stereo, mildenhall2021nerf} have been proposed that utilize multiple views of a scene to render novel views with geometric consistency. However, synthesizing novel views from a single image remains challenging due to the limited scene information available. Pixelsynth \cite{rockwell2021pixelsynth} and Geometry-Free View Synthesis \cite{rombach2021geometry} address this challenge by optimizing the synthesizing model using multi-view supervision. However, their generalization to in-the-wild scenes is hindered by the lack of large-scale multi-view datasets. Single-view MPI \cite{single_view_mpi} and MINE \cite{li2021mine} decompose the scene into multiple layers and utilize an inpainting network to extend each occluded layer. Additionally, AdaMPI \cite{han2022single} addresses complex 3D scene structures through a novel plane adjustment network. These MPI-based methods have demonstrated success in synthesizing realistic images, and thus, we adopt multiplane images as our basic synthesis tool. However, to the best of our knowledge, there are currently no publicly available methods for generating optical flow datasets from MPI. To extract optical flows from MPI, we propose a novel pipeline that differs from previous MPI-based image synthesis methods by utilizing layered depths and virtual camera poses. Additionally, to enhance the realism of the generated optical flow dataset, we introduce an independent object motion module for static and dynamic decoupling, as well as a depth-aware inpainting module to remove unnatural occupations.


\section{The Proposed MPI-Flow} 

In this section, we first briefly review the basics of our motivation and formulation for novel view image generation. Then we introduce the optical flow generation pipeline. Next, we present the details of two crucial components of our approach, including independent object motions and depth-aware inpainting.

\subsection{Motivation and Formulation}

Our goal is to generate a realistic novel view image $\textbf{I}_t\in\mathbb{R}^{H\times W\times 3}$ and the corresponding optical flow maps $\textbf{F}_{s\rightarrow t}\in\mathbb{R}^{H\times W\times 2}$ from single-view image $\textbf{I}_s\in\mathbb{R}^{H\times W\times 3}$. $H$ and $W$ are the height and width of the image, respectively. The two-dimensional array on the optical flow $\textbf{F}_{s\rightarrow t}$ represents the change of the corresponding pixel from image $\textbf{I}_s$ to image $\textbf{I}_t$. The input image, generated image, and optical flow together form a training pair.

To generate training pair, previous works \cite{aleotti2021learning, han2022realflow} wrap pixels from image $\textbf{I}_s$ to image $\textbf{I}_t$ with estimated flows. This inevitably leads to holes and artifacts in the image $\textbf{I}_t$, which damages the image realism. Recent work \cite{single_view_mpi, li2021mine, han2022single} on Multiplane Images (MPI) reveals that the layered depth representation of the single-view image can significantly improve the realism of the generated novel view image. 

We aim to tackle the image realism challenges in our methods and meanwhile enhance the optical flow realism and motion realism. Accordingly, we present an MPI-based optical flow dataset generation method, namely MPI-Flow. Figure \ref{fig:framework-MPI-Flow} shows the MPI-Flow framework for training pair generation. To construct MPI, given the input image $\textbf{I}_s$, an off-the-shelf monocular depth estimation network \cite{ranftl2020towards} is used to estimate its depth map. Then we use a neural network to construct $N$ fronto-parallel RGB$\sigma$ planes with color, density, and depth predicted by neural networks in the rays under novel viewpoints. 

To decouple dynamic objects, an instance segmentation network \cite{cheng2022masked} gives the object mask. Then we use bilinear interpolation to obtain the object masks under two viewpoints respectively. Using object masks and constructed MPI, we use volume rendering to render the separate novel view images, optical flow maps, and depths of dynamic objects and the static scene, respectively. The optical flow $\textbf{F}_{s\rightarrow t}$ can be obtained simply by adding the optical flows of the objects and the scene. However, due to different viewpoints, merging new images results in vacant areas and false occlusion. To this end, we design a depth-aware inpainting module, using rendered depths and object masks to fill the holes and repair false occlusion in the synthesized new image $\textbf{I}_{t}$.

\subsection{Optical Flow Data Generation}

\noindent \textbf{Novel View Image from MPI.} To render realistic image $\textbf{I}_t$ under a target viewpoint $\bm{\mu}_t$, we use pixel warping from the source-view MPI in a differentiable manner. Specifically, we use a neural network $\mathcal{F}$ as in \cite{han2022single} to construct $N$ fronto-parallel RGB$\sigma$ planes under source viewpoint $\bm{\mu}_{s}$ with color channels $\textbf{c}_n$, density channel $\bm{\sigma}_n$, and depth $\textbf{d}_n$ from the input image $\textbf{I}_s$ and its depth map $\textbf{D}_{s}$ as:
\begin{equation}
    \{(\textbf{c}_n, \bm{\sigma}_n, \textbf{d}_n)\}^N_{n=1} = \mathcal{F}(\textbf{I}_s, \textbf{D}_{s}),
\end{equation}
where $N$ is a predefined parameter that represents the number of planes in MPI. Each pixel ($x_t$, $y_t$) on the novel view image plane can be mapped to pixel ($x_s$, $y_s$) on $n$-th source MPI plane via homography function \cite{heyden2005multiple}:
\begin{equation}
    \left[x_{s}, y_{s}, 1\right]^{T} \sim \textbf{K}\left(\textbf{R}-\frac{\textbf{t} \textbf{n}^{T}}{\textbf{d}_{n}}\right) \textbf{K}^{-1}\left[x_{t}, y_{t}, 1\right]^{T},
    \label{eq:eq2}
\end{equation}
where $\textbf{R}$ and $\textbf{t}$ are the rotation and translation from the source viewpoints $\bm{\mu}_{s}$ to the target viewpoints $\bm{\mu}_{t}$, $\textbf{K}$ is the camera intrinsic, and $\textbf{n} = [0, 0, 1]$ is the normal vector. Thus, the color $\textbf{c}^{\prime}_{n}$ and density $\bm{\sigma}^{\prime}_{n}$ of each new plane for the novel view $\textbf{I}_t$ can be obtained via bilinear sampling. We use discrete intersection points between new planes and arbitrary rays passing through the scene and estimate integrals:
\begin{equation}
    \textbf{I}_t=\sum_{n=1}^{N}\left(\textbf{c}_{n}^{\prime} \bm{\alpha}_{n}^{\prime} \prod_{m=1}^{n-1}\left(1-\bm{\alpha}_{m}^{\prime}\right)\right), 
\label{eq:image}
\end{equation}
where $\bm{\alpha}^{\prime}_n=\exp \left(-\bm{\delta}_{n} \bm{\sigma}^{\prime}_{n}\right)$ and $\bm{\delta}_{n}$ is the distance map between plane $n$ and $n+1$ and we set the initial depth of MPI planes uniformly spaced in disparity as in \cite{han2022single}.

\

\noindent \textbf{Optical Flow from MPI.} Although MPI-based methods synthesize realistic images, reliable optical flow maps are also needed to train learning-based optical flow estimation models. Therefore, we propose adding an additional optical channel in each plane. To this end, we compute the optical flow on the $n$-th plane at pixel $[x_s, y_s]$ of source image $\textbf{I}_s$ by $\textbf{f}_{n}=[x_t-x_s, y_t-y_s]$ with a backward-warp process in terms of the inverse equivalent form of Equation \eqref{eq:eq2}:
\begin{equation}
    \left[x_{t}, y_{t}, 1\right]^{T} \sim \textbf{K}\left(\textbf{R}^{\dagger}-\frac{\textbf{t}^{\dagger} \textbf{n}^{T}}{\textbf{d}_{n}}\right) \textbf{K}^{-1}\left[x_{s}, y_{s}, 1\right]^{T},
\end{equation}
where $x_{s}$ and $y_{s}$ are uniformly sampled from a $H\times W$ grid. $\textbf{R}^{\dagger}$ and $\textbf{t}^{\dagger}$ are the inverses of $\textbf{R}$ and $\textbf{t}$, respectively.

To make sure that the optical flow maps match the novel view image $\textbf{I}_t$ perfectly, we propose to render $\textbf{F}_{s\rightarrow t}$ as in Equation \eqref{eq:image} in terms of volume rendering: 
\begin{equation}
    \textbf{F}_{s\rightarrow t}=\sum_{n=1}^{N}\left(\textbf{f}_n \bm{\alpha}_{n} \prod_{m=1}^{n-1}\left(1-\bm{\alpha}_{m}\right)\right), 
\label{eq:flow}
\end{equation}
where $\textbf{f}_n\in\mathbb{R}^{H\times W \times 2}$ is the optical flow maps on the $n$-th plane of image $\textbf{I}_s$. The pipeline implemented thus far models the optical flows resulting from camera motion without considering the potential presence of independently dynamic objects. However, real-world scenes are highly likely to contain such objects. Not incorporating their motions can lead to domain gaps by unrealistic optical flows.

\

\noindent \textbf{Independent Object Motions.} To model more realistic motions, we propose applying separate virtual motions to objects and static backgrounds extracted from the scene. Therefore, we utilize an instance segmentation network $\Omega$ \cite{cheng2022masked} for extracting the main object in the source image $\textbf{I}_s$ as:
\begin{equation}
    \textbf{M}_s = \Omega(\textbf{I}_s) \in \mathbb{R}^{H \times W},
\end{equation}
where $\textbf{M}_s$ is a binary mask to indicate the region of the object. To model the motions of the object $\textbf{M}_s$ in the scene, we construct separate viewpoints, including camera motion $\bm{\mu}_t^{sce}$ and object motion $\bm{\mu}_t^{obj}$. We then obtain the rendered scene novel view $\textbf{I}^{sce}_t$ and object novel view $\textbf{I}^{obj}_t$ as in Equation \eqref{eq:image}. The separate optical flows, $\textbf{F}^{sce}_{s\rightarrow t}$ and $\textbf{F}^{obj}_{s\rightarrow t}$ can also be obtained as in Equation \eqref{eq:flow}. The optical flows in $\textbf{F}_{s\rightarrow t}$ are mixed by the values in $\textbf{F}^{sce}_{s\rightarrow t}$ and $\textbf{F}^{obj}_{s\rightarrow t}$ in terms of mask $\textbf{M}_s$ to get the final optical flow maps containing camera motion and dynamic objects for training.

We can then use the bilinear interpolation to get the new object masks $\textbf{M}_t$ and $\textbf{M}_t^{obj}$ under the new viewpoints $\bm{\mu}_t$ and $\bm{\mu}_t^{obj}$ via bilinear sampling from $\textbf{M}_s$. Pixels in the new merged image are also selected according to the content masks $1-\textbf{M}_t$ and $\textbf{M}_t^{obj}$ from $\textbf{I}_t^{sce}$ and $\textbf{I}_t^{obj}$. Then, a simple inpainting strategy \cite{telea2004image} is used to fill the empty area in the new image with an inpainting mask calculated by $\textbf{M}_{fill}=\textbf{M}_t\odot(1-\textbf{M}_t^{obj})$.

\noindent \textbf{Depth-Aware Inpainting} Although merged images give a realistic visual effect, depth changes caused by camera motion and object motion can also cause unnatural occlusions. To solve this problem, we use volume rendering to obtain the depth $\textbf{D}_{t}^{sce}$ of the scene novel view:
\begin{equation}
    \textbf{D}_t=\sum_{n=1}^{N}\left(\textbf{d}_{n}^{\prime} \bm{\alpha}_{n}^{\prime} \prod_{m=1}^{n-1}\left(1-\bm{\alpha}_{m}^{\prime}\right)\right), 
\end{equation}
and the depth of the object novel view $\textbf{D}_{t}^{obj}$ can be obtained in the same way. We then utilize both depths to compute the occupation mask between the novel views:
\begin{equation}
    \textbf{M}_{occ}=(1-\textbf{M}_t) \odot \textbf{M}_t^{obj} \odot (\textbf{D}_{t}<\textbf{D}_{t}^{obj}),
\end{equation}
which indicates the background areas in front of the object. Therefore, we are able to restore the coincidence area between the object and the background in the new image $\textbf{I}_t$.

\begin{figure}[t]
   \centering
    \subcaptionbox{\label{11} Source Image}{\includegraphics[width = .95\linewidth]{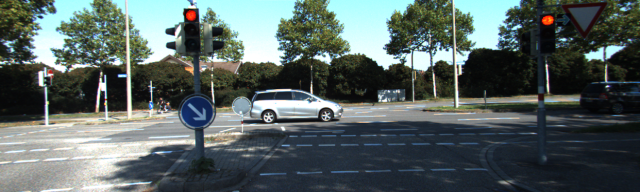}}\hfill
    \subcaptionbox{\label{22} + Camera Motion}{\includegraphics[width = .95\linewidth]{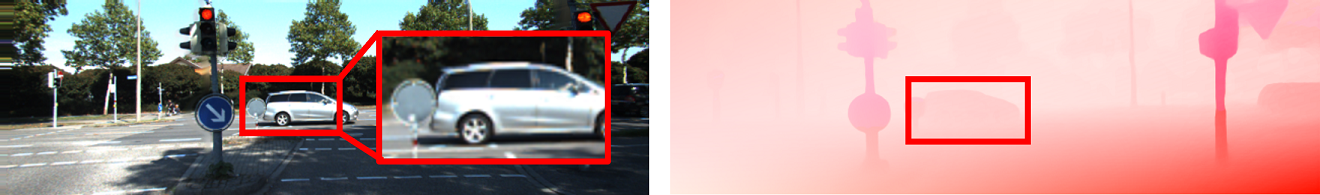}}\hfill
    \subcaptionbox{\label{33} + Independent Object Motion}{\includegraphics[width = .95\linewidth]{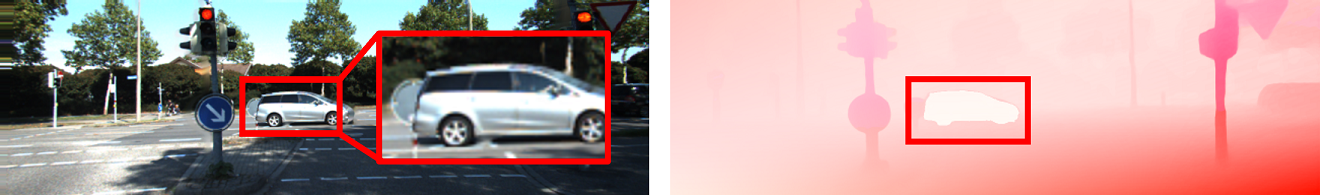}}\hfill
    \subcaptionbox{\label{44} + Depth-Aware Inpainting}{\includegraphics[width = .95\linewidth]{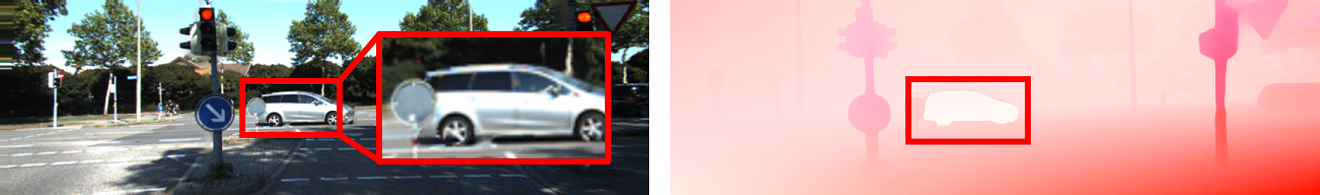}}
   \caption{Visualization of incrementally adding different modules to improve the realism of the generated data.}
   \label{fig:incremental}
\end{figure}

Figure \ref{fig:incremental} provides a detailed illustration of the incremental effects of MPI-Flow with and without independent object motion and depth-aware inpainting. Novel view images and optical flows from single-view images can be generated with MPI-Flow and only camera motion, as shown in Figures \ref{11} and \ref{22}. However, camera motion alone does not match the complex optical flows in real-world scenes. To address this issue, we introduce an independent object motion module, as shown in Figure \ref{33}, to ensure motion realism. To further enhance motion realism and address occlusion caused by object motion, we apply the depth-aware inpainting module, as shown in Figure \ref{44}.

\begin{table*}[t]  \small
\centering
\label{tab:ablation}
\begin{tabular}{c|c|cc|cc|cc|cc}
\hline
\rowcolor{graycolor} & ~ & \multicolumn{2}{c|}{Sintel.C} & \multicolumn{2}{c|}{Sintel.F} & \multicolumn{2}{c|}{KITTI 12} & \multicolumn{2}{c}{KITTI 15}  \\
\rowcolor{graycolor} \multirow{-2}{*}{Image Source} & \multirow{-2}{*}{Method} & EPE $\downarrow $ & $>$ 3 $\downarrow $ & EPE $\downarrow $ & $>$ 3 $\downarrow $ & EPE $\downarrow $ & Fl $\downarrow $ & EPE $\downarrow $ & Fl $\downarrow $ \\ \hline
\multirow{3}{*}{COCO} & Depthstillation \cite{aleotti2021learning} & \textbf{1.87} & 5.31 & 3.21 & 9.25 & 1.74 & 6.81 & 3.45 & 13.08 \\ 
& RealFlow \cite{han2022realflow} & N/A & N/A & N/A & N/A & N/A & N/A & N/A & N/A  \\ 
& MPI-Flow (ours) & \textbf{1.87} & \textbf{4.59} & \textbf{3.16} & \textbf{8.29} & \textbf{1.36} & \textbf{4.91} & \textbf{3.44} & \textbf{10.66}  \\ \hline
\multirow{3}{*}{DAVIS} & Depthstillation \cite{aleotti2021learning} & 2.70 & 7.52 & 3.81 & 12.06 & 1.81 & 6.89 & 3.79 & 13.22  \\
& RealFlow \cite{han2022realflow} & \textbf{1.73} & 4.81 & 3.47 & 8.71 & 1.59 & 6.08 & 3.55 & 12.52  \\
& MPI-Flow (ours) & 1.79 & \textbf{4.77} & \textbf{3.06} & \textbf{8.56} & \textbf{1.41} & \textbf{5.36} & \textbf{3.32} & \textbf{10.47}  \\ \hline
\multirow{3}{*}{KITTI 15 Test} & Depthstillation \cite{aleotti2021learning} & 4.02 & 9.08 & 4.96 & 13.23 & 1.77 & 5.97 & 3.99  & 13.34 \\ 
& RealFlow \cite{han2022realflow} & 3.73 & 7.36 & 5.53 & 11.31 & 1.27 & 5.16 & 2.43 & 8.86  \\
& MPI-Flow (ours) & \textbf{2.25} & \textbf{5.25} & \textbf{3.65} & \textbf{8.89} & \textbf{1.24} & \textbf{4.51} & \textbf{2.16} & \textbf{7.30}  \\ \hline
\multirow{3}{*}{KITTI 15 Train} & Depthstillation \cite{aleotti2021learning} & 2.84 & 7.18 & 4.31 & 11.24 & 1.67 & 5.71 & \{2.99\} & \{9.94\} \\
& RealFlow \cite{han2022realflow} & 4.06 & 7.68 & 4.78 & 11.44 & \textbf{1.25} & 5.02 & \{2.17\} & \{8.64\}  \\
& MPI-Flow (ours) & \textbf{2.41} & \textbf{5.39} & \textbf{3.82} & \textbf{9.11} & 1.26 & \textbf{4.66} & \{\textbf{1.88}\} & \{\textbf{7.16}\}  \\ \hline
\end{tabular}
\caption{The cross-dataset validation results and comparisons with other dataset generation methods from real images or videos are presented in this study. The "Image Source" column indicates the dataset used for optical flow training data generation. The evaluation results of RAFT trained on different datasets using different methods are reported. In cases where RealFlow fails to work on single-view images from COCO, the study indicates ``N/A". The curly braces ``\{\}" represent the use of the unlabeled evaluation set, which is the KITTI 15 training set in this table.}
\label{tab:dataset}
\end{table*}

\section{Experiments}


\subsection{Datasets}

\noindent \textbf{FlyingChairs \cite{dosovitskiy2015flownet} and FlyingThings3D \cite{ilg2017flownet}} are both popular synthetic datasets that train optical flow models. As a standard practice, we use ``Ch" and ``Th" respectively to represent the two datasets, and ``Ch$\rightarrow$Th" means training first on ``Ch" and fine-tuning on ``Th". By default, we use the RAFT pre-trained on ``Ch$\rightarrow$Th" to be fine-tuned on the generated datasets and evaluated on labeled datasets.

\noindent \textbf{COCO \cite{lin2014microsoft}} is a collection of single still images and ground truth with labels for object detection or panoptic segmentation tasks. We sample 20k single-view still images from the train2017 split following Depthstillation \cite{aleotti2021learning} to generate virtual images and optical flow maps.

\noindent \textbf{DAVIS} \cite{perazzi2016benchmark} provides high-resolution videos and it is widely used for video object segmentation. We use all the 10581 images of the unsupervised 2019 challenge to generate datasets by MPI-Flow and other state-of-the-art optical flow dataset generation methods.

\noindent \textbf{KITTI2012 \cite{geiger2012we} and KITTI2015 \cite{menze2015object}} are well-known benchmarks for optical flow estimation. There are multi-view extensions (4,000 for training and 3,989 for testing) datasets with no ground truth. We use the multi-view extension images (training and testing) of KITTI 2015 to generate datasets, separately. By default, we evaluate the trained models on KITTI 12 training set and KITTI 15 training set in the tables following \cite{aleotti2021learning} and \cite{han2022realflow}, abbreviated as ``KITTI 15" and ``KITTI 12" in the following tables.

\noindent \textbf{Sintel \cite{butler2012naturalistic}} is derived from the open-source 3D animated short film Sintel. The dataset has 23 different scenes. The stereo images are RGB, while the disparity is grayscale. Although not a real-world dataset, we use it to verify the model's generalization across domains.


\subsection{Implementation Details}
Firstly, we provide a description of the learning-based optical flow estimation models that were utilized in our experiments. Subsequently, we outline the experimental parameters and setup along with the evaluation formulation.

\noindent \textbf{Optical Flow networks.} To evaluate how effective our generated data are at training optical flow models, we select RAFT \cite{teed2020raft}, which represents state-of-the-art architecture for supervised optical flow and has excellent generalization capability. By default, we train RAFT on generated data for 200K steps with a learning rate of $1\times10^{-4}$ and weight decay of $1\times10^{-5}$, batch size of $6$, and $288\times960$ image crops. This configuration is the default setting of RAFT fitting on KITTI with two GPUs but four times the number of training steps, following \cite{han2022realflow}. For the rest of the setup, we use the official implementation of RAFT without any modifications. All evaluations are performed on a single NVIDIA GeForce RTX 3090 GPU \footnote{RAFT with the same parameters loaded on different GPUs yield slightly different evaluation results. Therefore, to ensure fair comparison, we download the official model weights with the best performance provided by the compared methods and evaluate them on the same GPU.}.

\noindent \textbf{Virtual Camera Motion.} To generate the novel view images from COCO and DAVIS, we adopt the same settings in \cite{aleotti2021learning} to build the virtual camera. For KITTI, we empirically build the camera motion with three scalars where $t_x, t_y$ are in $[-0.2, 0.2]$ and $t_z$ are in $[0.1, 0.35]$. We build the camera rotation with three Euler angles $a_x, a_y, a_z$ in $[-\frac{\pi}{90}, \frac{\pi}{90}]$. We use single camera motion for each image from COCO but multiple camera motions ($\times4$ by default) from DAVIS and KITTI as in \cite{han2022realflow}, due to the small number of images and the homogeneity of the scene in video data. We show how the number of camera motions impacts the optical flow network performance in the discussion.

\noindent \textbf{Evaluation Metrics.} We report evaluation results on the average End-Point Error (EPE) and two error rates, respectively the percentage of pixels with an absolute error greater than 3 ($>$ 3) or both absolute and relative errors greater than 3 and 5\% respectively (Fl) on all pixels.

\subsection{Comparison with State-of-the-art Methods}

In this section, we evaluate the effectiveness of the MPI-Flow generation pipeline on public benchmarks. We will highlight the best results in \textbf{bold} and \underline{underline} the second-best if necessary among methods trained in fair conditions.

\begin{table}[t]  \small
\centering
\label{tab:ablation}
\begin{tabular}{c|cc|cc}
\hline
\rowcolor{graycolor} & \multicolumn{2}{c|}{KITTI 12} & \multicolumn{2}{c}{KITTI 15} \\
\rowcolor{graycolor} \multirow{-2}{*}{Dataset} & EPE $\downarrow $ & Fl $\downarrow $ & EPE $\downarrow $ & Fl $\downarrow $ \\ \hline
 Ch$\rightarrow$Th \cite{ilg2017flownet} & 2.08 & 8.86 & 5.00 & 17.44 \\
 dDAVIS \cite{aleotti2021learning} & 1.81 & 6.89 & 3.79 & 13.22  \\
 MF-DAVIS & \underline{1.61} & \underline{6.41} & \underline{3.77} & \underline{12.40}  \\ 
 dCOCO \cite{aleotti2021learning} & 1.80 & 6.66 & 3.80 & 12.44  \\
 MF-COCO & \textbf{1.59} & \textbf{6.22} & \textbf{3.68} & \textbf{11.95}  \\ \hline
\end{tabular}
\caption{Results on RAFT trained from scratch under the same setting. ``dX'' are from Depthstillation \cite{aleotti2021learning} while ``MF-X'' are from our proposed MPI-Flow}
\label{tab:without-pretrain}
\end{table}

\begin{figure*}[t]
   \centering
    \subcaptionbox{\label{sub-baseline} Source Image}{\includegraphics[width = .24\linewidth]{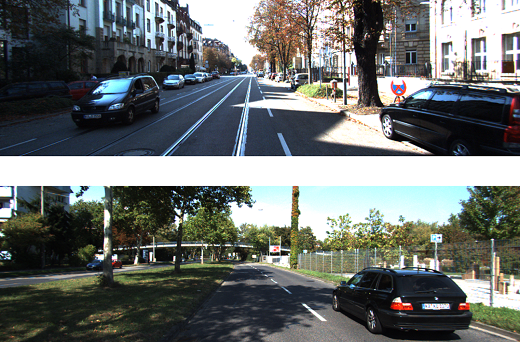}}\hfill
    \subcaptionbox{\label{sub-ds} Depthstillation}{\includegraphics[width = .24\linewidth]{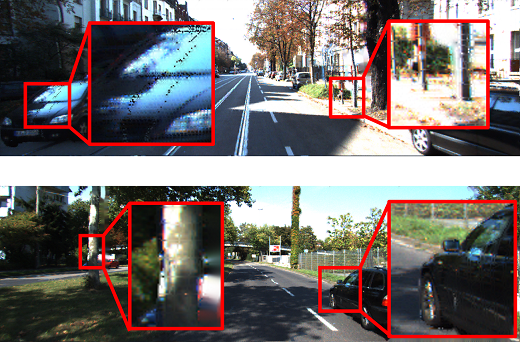}}\hfill
    \subcaptionbox{\label{sub-baseline2} RealFlow}{\includegraphics[width = .24\linewidth]{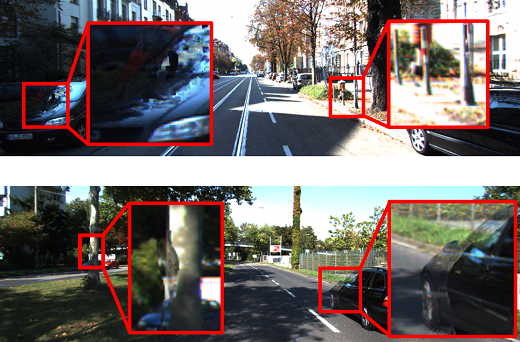}}
    \subcaptionbox{\label{sub-ours} MPI-Flow (ours)}{\includegraphics[width = .24\linewidth]{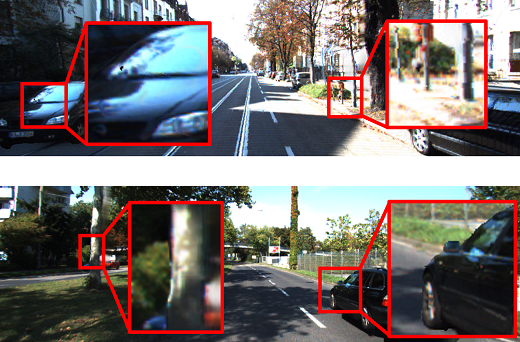}}
   \caption{Qualitative results of generated images and optical flows from the KITTI 2015 training set. (a) contains the input source images. (b), (c), and (d) contain generated images of RealFlow \cite{han2022realflow}, Depthstillation \cite{aleotti2021learning}, and our proposed MPI-Flow. MPI-Flow eliminates the artifact in the new image and guarantees image realism. (Best viewed with zoom-in)}
   \label{fig:sync-image}
\end{figure*}

\begin{figure}[t]
   \centering
    \subcaptionbox{\label{1st} Source Image from KITTI 15}{\includegraphics[width = .99\linewidth]{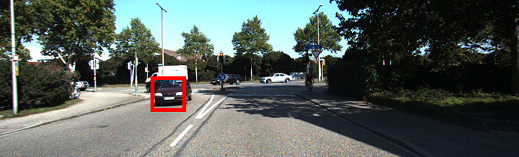}}
    \subcaptionbox{\label{1st} Depthstillation}{\includegraphics[width = .32\linewidth]{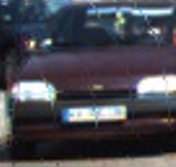}}
    \subcaptionbox{\label{8th} RealFlow}{\includegraphics[width = .32\linewidth]{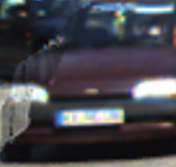}}
    \subcaptionbox{\label{16th} MPI-Flow (ours)}{\includegraphics[width = .32\linewidth]{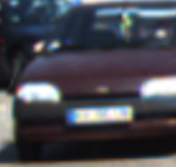}}
   \caption{Visualization of generated images with Depthstillation, RealFlow, and our proposed MPI-Flow from KITTI. Note that RealFlow generates such data using two frames with estimated optical flow.}
   \label{fig:samples-real}
\end{figure}

\begin{figure}[t]
   \centering
    \subcaptionbox{\label{sub-baseline} Depthstillation}{\includegraphics[width = .46\linewidth]{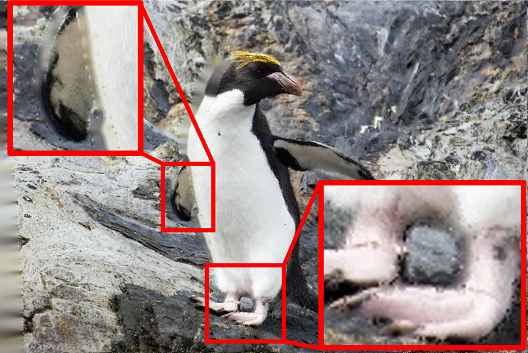}}\hfill
    \subcaptionbox{\label{sub-ours} MPI-Flow (ours)}{\includegraphics[width = .46\linewidth]{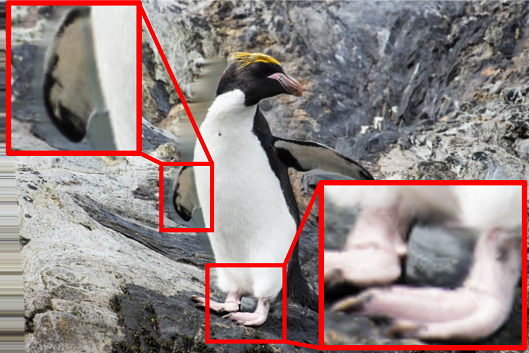}}
   \caption{Visualization of generated images with Depthstillation and our proposed MPI-Flow using a single view.}
   \label{fig:single-image}
\end{figure}

\noindent \textbf{Comparison with Dataset Generation Methods.} As a method for generating datasets from real-world images, we compare MPI-Flow with Depthstillation \cite{aleotti2021learning} and RealFlow \cite{han2022realflow}, which are representative works in real-world dataset generation. In order to evaluate the effectiveness of MPI-Flow, we follow the procedures of Depthstillation and RealFlow to construct training sets from four different datasets, including COCO, DAVIS, KITTI 15 multi-view train set, and KITTI 15 multi-view test set. To ensure fair comparisons, we conduct experiments with a similar number of training sets as our competitors. Specifically, for DAVIS and KITTI, we generate four motions for each image to match RealFlow, which trains RAFT for four EM iterations with four times the amount of data. For COCO, we follow the exact same setup as Depthstillation. Since Depthstillation does not provide details for KITTI 15 train, we use its default settings to obtain the results. Trained models are evaluated on the training sets of Sintel, KITTI 12, and KITTI 15. We report the evaluation results of models with the best performance in the paper for Depthstillation and RealFlow. Furthermore, we conduct cross-dataset experiments where RAFT is trained with one generated dataset and evaluated with another. 

As shown in Table \ref{tab:dataset}, even with the same amount of data, our approach gains significant improvements and generalization over multiple datasets. When trained and tested with the same KITTI 15 Train image source, our EPE outperforms the second-best by a remarkable 0.29. When trained and tested with different image sources, our MPI-Flow demonstrates clear improvements over the competitors on almost all the evaluation settings. Notably, MPI-Flow achieves much better performance, even though RealFlow requires two consecutive frames to generate datasets, while MPI-Flow needs only one still image.

It is worth comparing MPI-Flow with Depthstillation under the same settings to evaluate their performance. For this comparison, we use the exact same settings as Depthstillation. Specifically, we generate MPI-Flow datasets with 1) no object motion, 2) the same camera parameters, 3) one camera motion per image, and 4) without pre-training on Ch$\rightarrow$Th. The datasets generated are named MF-DAVIS and MF-COCO, while dCOCO and dDAVIS are generated from Depthstillation. The models are trained from scratch with the respective datasets. RealFlow is not included in this table as it requires a pre-trained optical flow model to generate datasets. The evaluation results are shown in Table \ref{tab:without-pretrain}. Our method outperforms Depthstillation with significant improvements on both COCO and DAVIS, demonstrating the importance of image realism.

\begin{table}[t]  \small
\centering
\label{tab:ablation}
\begin{tabular}{c|cc|cc}
\hline
\rowcolor{graycolor} & \multicolumn{2}{c|}{KITTI 12} & \multicolumn{2}{c}{KITTI 15} \\ 
 \rowcolor{graycolor} \multirow{-2}{*}{Method} & EPE $\downarrow $ & Fl $\downarrow $ & EPE $\downarrow $ & Fl $\downarrow $ \\ \hline
 SemiFlowGAN \cite{lai2017semi} & - & - & \{16.02\} & \{38.77\} \\ 
 FlowSupervisor \cite{im2022semi} & - & - & \{3.35\} & \{11.12\} \\ 
 DistractFlow \cite{jeong2023distractflow} & - & - & \{3.01\} & \{11.7\} \\ 
 Meta-Learning \cite{min2023meta} & - & - & \{2.81\} & - \\ 
 SimFlow \cite{im2020unsupervised} & - & - & \{5.19\}  & - \\
 ARFlow \cite{liu2020learning}        & 1.44 & - & \{2.85\}  & - \\
 UFlow \cite{jonschkowski2020matters} & 1.68 & - & 2.71 & 9.05 \\ 
 UpFlow \cite{luo2021upflow}          & 1.27 & - & 2.45 & - \\
 SMURF \cite{stone2021smurf}          & - & - & \{2.00\} & \{\textbf{6.42}\} \\
 MPI-Flow & \textbf{1.18} & \textbf{4.46} & \{\textbf{1.80}\} & \{6.63\}  \\ \hline 
\end{tabular}
\caption{Comparison with semi-supervised and unsupervised methods. ‘-’ indicates no results reported.}
\label{tab:unsupervised}
\end{table}

\begin{table*}[t]  \small
\centering
\label{tab:ablation}
\begin{tabular}{ccc|cc|cc|cc|cc}
\hline
 \rowcolor{graycolor} Dynamic & Depth-Aware & Multiple & \multicolumn{2}{c|}{Sintel.C} & \multicolumn{2}{c|}{Sintel.F} & \multicolumn{2}{c|}{KITTI 12} & \multicolumn{2}{c}{KITTI 15} \\
 \rowcolor{graycolor} Objects & Inpainting & Objects & EPE $\downarrow $ & $>$ 3 $\downarrow $ & EPE $\downarrow $ & $>$ 3 $\downarrow $ & EPE $\downarrow $ & Fl $\downarrow $ & EPE $\downarrow $ & Fl $\downarrow $ \\ \hline
  \ding{53} & \ding{53} & \ding{53} & 2.58 & 6.10 & 4.04 & 9.67 & \textbf{1.20} & \textbf{4.34} & \{2.46\} & \{8.38\} \\
  \ding{51} & \ding{53} & \ding{53} & 2.41 & 5.39 & 3.82 & 9.11 & 1.26 & 4.66 & \{\textbf{1.88}\} & \{7.16\} \\ 
  \ding{51} & \ding{51} & \ding{53} & 2.37 & 5.44 & 3.71 & 9.05 & 1.25 & 4.58 & \{1.91\} & \{7.06\}  \\
  \ding{51} & \ding{51} & \ding{51} & \textbf{2.20} & \textbf{5.35} & \textbf{3.67} & \textbf{9.03} & 1.23 & 4.46 & \{1.92\} & \{\textbf{7.05}\}  \\ \hline
\end{tabular}
\caption{Ablation experiments. Settings used are marked with a checkmark. Here we only perform four camera motions per image for these experiments due to the limitation of computational resources.}
\label{tab:ablations}
\end{table*}

\begin{table}[t]  \small
\centering
\label{tab:ablation}
\begin{tabular}{c|cc|c}
\hline
 \rowcolor{graycolor} & \multicolumn{2}{c|}{KITTI 15} & KITTI 15 test \\ 
 \rowcolor{graycolor} \multirow{-2}{*}{Method} & EPE $\downarrow $ & Fl $\downarrow $ & Fl $\downarrow $ \\ \hline
 PWC-Net \cite{sun2018pwc} & \{2.16\} & \{9.80\} & 9.60  \\
 LiteFlowNet \cite{hui2018liteflownet} & \{1.62\} & \{5.58\} & 9.38 \\
 IRR-PWC \cite{hur2019iterative} & \{1.63\} & \{5.32\} & 7.65 \\
 RAFT \cite{teed2020raft} & \{0.63\} & \{1.50\} & 5.10  \\
 Depthstillation \cite{han2022realflow} & - & - & - \\
 AutoFlow \cite{sun2021autoflow} & - & - & 4.78  \\
 RealFlow \cite{han2022realflow} & \{\textbf{0.58}\} & \{1.35\} & 4.63  \\ 
 MPI-Flow & \{\textbf{0.58}\} & \{\textbf{1.30}\} & \textbf{4.58} \\ \hline
\end{tabular}
\caption{Comparison with supervised methods fine-tuned or trained on KITTI 15 train set.}
\label{tab:supervised}
\end{table}   

\noindent \textbf{Comparison with Unsupervised Methods.} Another way to utilize real-world data is through unsupervised learning, which learns optical flow pixels directly without the need for optical flow labels. In order to further demonstrate the effectiveness of MPI-Flow, we compare our method with the existing literature on unsupervised methods. The results of this comparison can be seen in Table \ref{tab:unsupervised}. All methods are evaluated under the condition that only images from the evaluation set could be used, without access to ground truth labels. For evaluation, we train the RAFT on our generated dataset with images from the KITTI 15 training set. Our MPI-Flow outperform all unsupervised methods in terms of EPE on both the KITTI 12 training set and KITTI 15 training set, with no need for any unsupervised constraints. However, our method performs better on EPE but has slightly lower Fl than SMURF, mainly because SMURF employs multiple frames for training.

\noindent \textbf{Comparison with Supervised Methods.} To further prove the effectiveness of MPI-Flow, we use KITTI 15 train set to fine-tune RAFT pre-trained by our generated dataset with images from KITTI 15 test set. The evaluation results on KITTI 15 train and KITTI 15 test are shown in Table \ref{tab:supervised}. We achieve state-of-art performance on KITTI 2015 test benchmark compared to supervised methods on training RAFT.

\noindent \textbf{Qualitative Results.} Figure \ref{fig:sync-image} show the generated images from the methods utilizing real-world images, as presented in Table \ref{tab:dataset}. In this comparison, we use images from the KITTI 15 dataset as source image input. The images generated by RealFlow \cite{han2022realflow} and Depthstillation \cite{aleotti2021learning} with artifacts degrade the image realism. In contrast, MPI-Flow generates more realistic images than the other two methods. More results are shown in Figures \ref{fig:samples-real} and \ref{fig:single-image}.

\subsection{Discussion}

To verify the effectiveness of the proposed MPI-Flow, we discuss the performance of models trained with generated datasets with different settings. We show more discussion and evaluation results in the supplementary material.

\noindent \textbf{Object Motion and Depth-Aware Inpainting.} We conduct a series of ablation studies to analyze the impact of different choices in our proposed MPI-Flow for new image synthesis, including object motion, depth-aware inpainting, and multiple objects. ``Multiple objects'' indicates multiple moving objects in each image. To measure the impact of these factors, we generate new images from the KITTI 15 training set to train RAFT and evaluate them on the KITTI 12 training set and KITTI 15 training set. Because there are multiple combinations of these factors, we test by incrementally adding components of our approach, as shown in Table \ref{tab:ablations}. In the first row, we show the performance achieved by generating new images without dynamic objects, thus assuming that optical flow comes from camera motion only. Then we incrementally add single-object motion, depth-aware inpainting, and multi-object motion to model a more realistic scenario. There are considerable improvements in almost all datasets on various metrics, except for the KITTI 12 training set, possibly due to the infrequent dynamic object motion in this dataset. The EPE on KITTI 15 remains relatively stable within the normal margin after adding the depth-aware inpainting module and the multi-object trick.

\begin{table}[t]  \small
\centering
\label{tab:ablation}
\begin{tabular}{c|cc|cc}
\hline
 \rowcolor{graycolor} Motions & \multicolumn{2}{c|}{KITTI 12} & \multicolumn{2}{c}{KITTI 15} \\
 \rowcolor{graycolor} Per Image & EPE $\downarrow $ & Fl $\downarrow $ & EPE $\downarrow $ & Fl $\downarrow $ \\ \hline
1  & 1.29 & 4.58 & 2.02 &  7.35 \\
4  & 1.23 & 4.46 & 1.92 & 7.05 \\ 
10 & 1.21 & 4.49 & 1.82 & 7.02 \\
20 & 1.20 & \textbf{4.41} & \textbf{1.79} & 6.86 \\ 
40 & \textbf{1.18} & 4.46 & 1.80 & \textbf{6.63} \\ \hline
\end{tabular}
\caption{Effect of amount of virtual camera motions (training pairs) per source image.}
\label{tab:num-of-motion-kitti}
\end{table}

\begin{table}[t]  \small
\centering
\label{tab:ablation}
\begin{tabular}{c|cc|cc}
\hline
 \rowcolor{graycolor} Objects & \multicolumn{2}{c|}{KITTI 12} & \multicolumn{2}{c}{KITTI 15} \\
 \rowcolor{graycolor} Per Image & EPE $\downarrow $ & Fl $\downarrow $ & EPE $\downarrow $ & Fl $\downarrow $ \\ \hline
1 & 1.25 & 4.58 & \textbf{1.91} & 7.06 \\
2 & 1.26 & 4.76 & 2.02 & 7.19 \\ 
4 & \textbf{1.23} & \textbf{4.46} & 1.92 & \textbf{7.05} \\ \hline
\end{tabular}
\caption{Effect of amount of dynamic objects per image.}
\label{tab:num-of-objects-kitti}
\end{table}

\noindent \textbf{Camera Motion Parameters.} We also conduct a series of ablation studies to analyze the impact of different parameters on camera motions. We first show the effect of $t_z$, which indicates the range of moving forward and backward distances, as shown in Table \ref{tab:ablations}. We set the minimum $t_z$ to $0.1$ by default, considering that most cameras on vehicles only move forward in the KITTI dataset. We also show the effect of $t_x$ and $t_y$, representing the distance from left to right and from up to down, respectively. Because there are multiple combinations of parameters, we only test a specific parameter of our method in isolation. Settings used by default are \underline{underlined}. The results show that a more reasonable range of camera motion leads to better performance.

\begin{table}[t]  \small
\centering
\begin{tabular}{c|cc|cc}
\hline
 \rowcolor{graycolor} & \multicolumn{2}{c|}{KITTI 12} & \multicolumn{2}{c}{KITTI 15} \\
 \rowcolor{graycolor} \multirow{-2}{*}{$t_z$} & EPE $\downarrow $ & Fl $\downarrow $ & EPE $\downarrow $ & Fl $\downarrow $ \\ \hline
$0.1\sim0.45$ & 1.25 & 4.60 & \{1.79\} & \{6.68\} \\
\underline{$0.1\sim0.35$} & 1.18 & 4.46 & \textbf{\{1.80\}} & \textbf{\{6.63\}} \\
$0.1\sim0.25$ & 1.19 & 4.50 & \{1.86\} & \{6.71\} \\ 
$0.0\sim0.35$ & \textbf{1.12} & \textbf{4.32} & \{1.94\} & \{6.92\} \\ \hline
\end{tabular}
\caption{Effect of motion parameters $t_z$.}
\label{tab:ab-tz}
\end{table}

\begin{table}[t]  \small
\centering
\begin{tabular}{c|cc|cc}
\hline
 \rowcolor{graycolor} & \multicolumn{2}{c|}{KITTI 12} & \multicolumn{2}{c}{KITTI 15} \\
 \rowcolor{graycolor} \multirow{-2}{*}{$t_x$ and $t_y$} & EPE $\downarrow $ & Fl $\downarrow $ & EPE $\downarrow $ & Fl $\downarrow $ \\ \hline
$-0.3\sim0.3$ & 1.22 & \textbf{4.42} & \{1.92\} & \{6.83\} \\
\underline{$-0.2\sim0.2$} & \textbf{1.18} & 4.46 & \textbf{\{1.80\}} & \textbf{\{6.63\}} \\ 
$-0.1\sim0.1$ & 1.32 & 4.99 & \{1.98\} & \{7.02\} \\ \hline
\end{tabular}
\caption{Effect of motion parameters $t_x$ and $t_y$.}
\label{tab:ab-txy}
\end{table}

\noindent \textbf{Model Architectures} The default model used in our paper is RFAT \cite{teed2020raft}. Table \ref{tab:pwc} shows that our proposed MPI-Flow also works for PWC-Net \cite{sun2018pwc} compared with depthstillation \cite{aleotti2021learning} in a fair setting. We generate data from COCO and DAVIS and train PWC-Net, in which the results are still better than PWC-Net trained on Ch$\rightarrow$Th or datasets generated from depthstillation. 

\begin{table}[t]  \small
\centering
\begin{tabular}{c|cc|cc}
\hline
\rowcolor{graycolor} & \multicolumn{2}{c|}{KITTI 12} & \multicolumn{2}{c}{KITTI 15} \\
\rowcolor{graycolor} \multirow{-2}{*}{Dataset} & EPE $\downarrow $ & Fl $\downarrow $ & EPE $\downarrow $ & Fl $\downarrow $ \\ \hline
 Ch$\rightarrow$Th \cite{ilg2017flownet} & 4.14 & 21.38 & 10.35 & 33.67 \\ \cline{2-5}
 dDAVIS \cite{aleotti2021learning} & 2.81 & 11.29 & \textbf{6.88} & 21.87 \\
 MF-DAVIS & \textbf{2.70} & \textbf{11.25} & 6.92 & \textbf{20.97} \\ \cline{2-5}
 dCOCO \cite{aleotti2021learning} & 3.16 & 13.30 & 8.49 & 26.06 \\
 MF-COCO & \textbf{3.02} & \textbf{11.20} & \textbf{8.22} & \textbf{23.40} \\ \hline
\end{tabular}
\caption{Results on PWC-Net trained on generated datasets from scratch under the same setting.}
\label{tab:pwc}
\end{table}

\noindent \textbf{Amount of Virtual Motions.} We can generate multiple camera motions with multiple dynamic objects for any given single image and thus a variety of paired images and ground-truth optical flow maps. Thus we can increase the number of camera motions to generate more data. Table \ref{tab:num-of-motion-kitti} shows the impact of different amounts of camera motions on model performance. Interestingly, MPI-Flow with $4$ motions per image already allows for strong generalization to real domains, outperforming the results achieved using synthetic datasets shown in the previous evaluation results. Increasing the motions per image by factors $10$ and $20$ both lead to better performance on KITTI 12 and KITTI 15 compared to $4$ and $1$. Using $40$ motions per image gives the best performance on KITTI 15 in terms of Fl. It indicates that a more variegate image content in the generated dataset may be beneficial for generalization to real applications. Table \ref{tab:num-of-objects-kitti} shows the effect of amounts of dynamic objects on model performance. Increasing the number of dynamic objects improves the performance of the model on KITTI 12 but slightly decreases it on KITTI 15 in terms of EPE. 

\noindent \textbf{Quantity of Source Images} The number of source images affects the scene diversity of the generated dataset. Empirically, more source images will be more conducive to model learning, as verified in Table \ref{tab:coco-source}. We verify the effect of the number of source images on the MF-COCO. The model performance is significantly improved by increasing the number of source images.

\begin{table}[t]  \small
\centering
\begin{tabular}{c|c|cc|cc}
\hline
\rowcolor{graycolor} & ~ & \multicolumn{2}{c|}{KITTI 12} & \multicolumn{2}{c}{KITTI 15} \\
 \rowcolor{graycolor} \multirow{-2}{*}{Dataset}& \multirow{-2}{*}{Quantity} & EPE $\downarrow $ & Fl $\downarrow $ & EPE $\downarrow $ & Fl $\downarrow $ \\ \hline
 Ch$\rightarrow$Th \cite{ilg2017flownet} & 47K & 2.08 & 8.86 & 5.00 & 17.44  \\
 MF-COCO & 20K & 1.59 & 6.22 & 3.68 & 11.95  \\
 MF-COCO & 120K & \textbf{1.51} & \textbf{5.94} & \textbf{3.41} & \textbf{11.16}  \\ \hline
\end{tabular}
\caption{Effect of amount of source images.}
\label{tab:coco-source}
\end{table}

\section{Conclusion}

In this paper, we present a new framework for generating optical flow datasets, which addresses two main challenges: image realism and motion realism. Firstly, we propose an MPI-based image rendering pipeline that generates realistic images with corresponding optical flows from novel viewpoints. This pipeline utilizes volume rendering to address image artifacts and holes, leading to more realistic images. Secondly, we introduce an independent object motion module that separates dynamic objects from the static scene. By decoupling object motion, we further improve motion realism. Additionally, we design a depth-aware inpainting module that handles unnatural occlusions caused by object motion in the generated images. Through these novel designs, our approach achieves superior performance on real-world datasets compared to both unsupervised and supervised methods for training learning-based models.

\vspace{1mm}
\noindent\textbf{Acknowledgments}
This work was supported by the National Natural Science Foundation of China (12202048, 62171038, and 62171042), the R\&D Program of Beijing Municipal Education Commission (KZ202211417048), and the Fundamental Research Funds for the Central Universities.


\end{document}